\documentclass[journal]{IEEEtran}
\usepackage{bm}
\usepackage{amsmath}
\usepackage{amssymb}
\usepackage{amsthm}
\usepackage{mathrsfs}
\usepackage{enumerate}
\usepackage{multirow}
\usepackage{color}
\usepackage{subfigure}
\usepackage[square, comma, sort&compress, numbers]{natbib}
\usepackage{times}
\usepackage{breakurl}
\usepackage{array}
\usepackage{verbatim}
\usepackage{algorithm}
\usepackage{algpseudocode}
\usepackage[colorlinks,linkcolor=red]{hyperref}
\definecolor{orange}{rgb}{1,0.5,0} %v2
\definecolor{purple}{rgb}{0.5,0,0.5} %v3

\makeatletter
\def\hlinewd#1{%
  \noalign{\ifnum0=`}\fi\hrule \@height #1 \futurelet
   \reserved@a\@xhline}
\makeatother

\ifCLASSINFOpdf
  \usepackage[pdftex]{graphicx}
  \usepackage{graphics}
  \usepackage{graphicx}
  \usepackage{epsfig}
  \usepackage{epstopdf}
  \usepackage{authblk}
  \graphicspath{{./fig/}}
  \DeclareGraphicsExtensions{.pdf,.jpeg,.png,.eps}
\else
  \usepackage[dvips]{graphicx}
  \graphicspath{{./fig/}}
  \DeclareGraphicsExtensions{.eps}
\fi

\begin{document}

\title{A Competitive Method to VIPriors Object Detection Challenge}
%\author{Fei Shen, Xin He, Mengwan Wei , Yi Xie}
\author[*]{Fei Shen}
\author[+]{Xin He}
\author[*]{Mengwan Wei}
\author[*]{Yi Xie}
\affil[*]{Huaqiao University, feishen@hqu.edu.cn}
\affil[+]{Wuhan University of Technology }
\markboth{}%
%\markboth{Journal of \LaTeX\ Class Files,~Vol.~6, No.~1, January~2007}%
{Shell \MakeLowercase{\textit{et al.}}: Bare Demo of IEEEtran.cls for Journals}

% make the title area
\maketitle

\begin{abstract}
In this report, we introduce the technical details of our submission to the VIPriors object detection challenge.
Our solution is based on mmdetction of a strong baseline open-source detection toolbox.
Firstly, we introduce an effective data augmentation method to address the lack of data problem, which contains bbox-jitter, grid-mask, and mix-up.
Secondly, we present a robust region of interest (ROI) extraction method to learn more significant ROI features via embedding global context features.
Thirdly, we propose a multi-model integration strategy to refinement the prediction box, which weighted boxes fusion (WBF).
Experimental results demonstrate that our approach can significantly improve the average precision (AP) of object detection on the subset of the COCO2017 dataset.
Code is available at \url{https://github.com/muzishen/VIPriors-Object-Detection-Challenge}.
\end{abstract}

%\begin{IEEEkeywords}
%Deep Learning, Graph Network, Spatial Significance, Vehicle Re-identification
%\end{IEEEkeywords}

\IEEEpeerreviewmaketitle

%-------------------------------------------------
\section{Introduction}
VIPriors object detection challenge is a workshop in the ECCV2020 conference.
It focuses on obtaining high average precision (AP) on a subset of the COCO2017 object detection dataset.
%The subset of COCO2017 consists of 5873 train images, 4946 val images and 4952 test images.
It is very challenging for object detection due to the adverse influence of the sample class imbalance and lacking of data.
To address these challenges, we focus on the data augmentation in data preprocess and adapt the bbox-jitter \footnote{https://github.com/cizhenshi}, grid-mask \cite{gridmask}, and mix-up \cite{mixup} methods to improve the diversity of object images.
Then, to make the detector to learn more prominent features, we closely connect the global context features to the ROI features at each stage.
 To overcome the lack of data, we respectively trained three moderate backbones network of Cascade-RCNN \cite{cascade}.
 Finally, we adopt to weighted boxes fusion (WBF) \cite{wbf} method of the multiple-model ensemble on the test dataset.
Experimental results demonstrate our approach can significantly improve the object detection performance and achieve a competitive result on the test set.
According to the rules of the competition, we do not use any external image/video data or pre-trained weights.
The implementation details of the above are described in section 2 and section 3.

\begin{figure}[htp]
	\centering
	\includegraphics[width=1\linewidth]{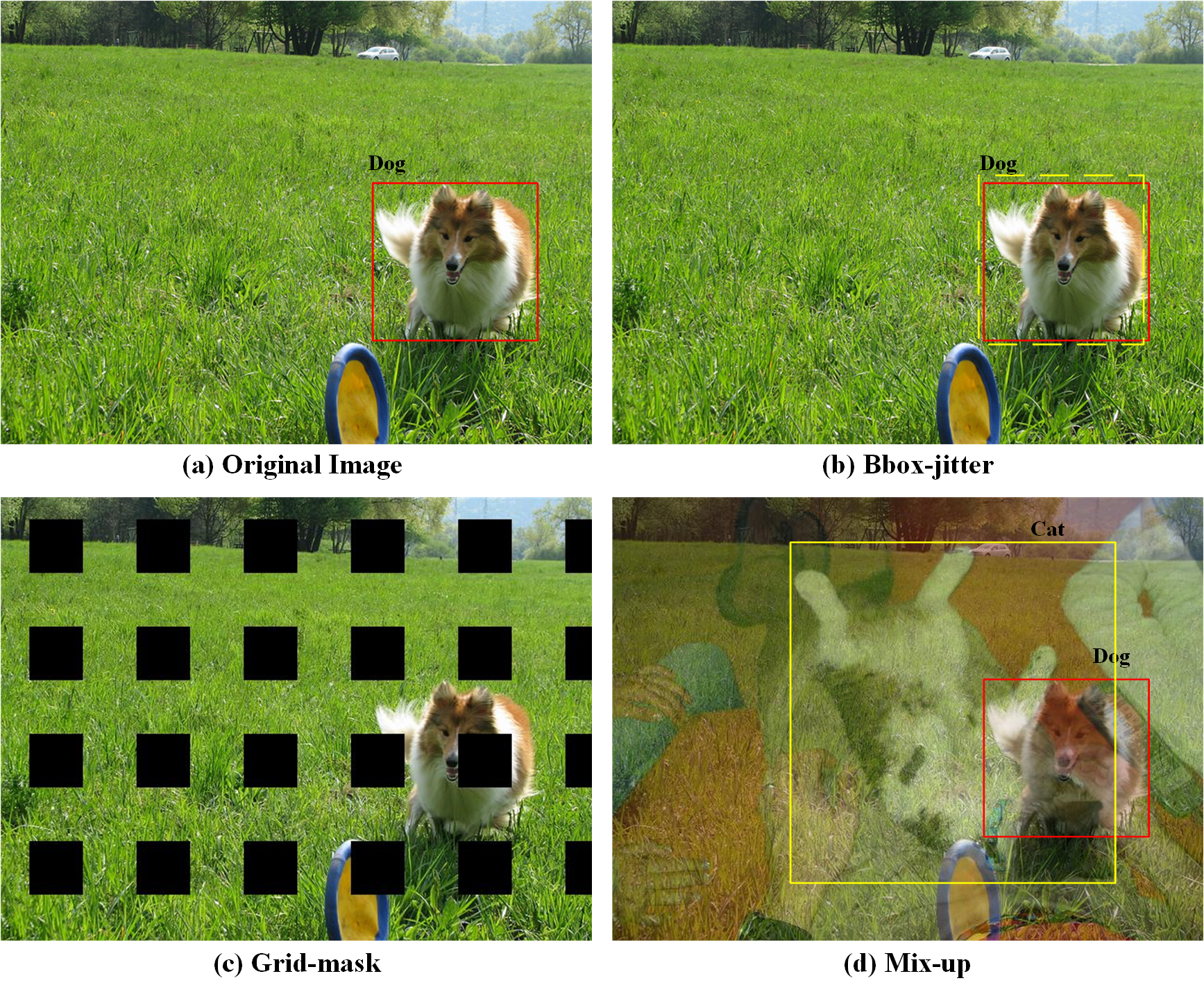}
\vspace{-0.5cm}
	\caption{Visualization of three data augmentation methods. (a) The red boxes are the detection result of the original image. (b) The yellow boxes are the result of the original image detection result via using Bbox-jitter. (c) The image generated from the original image using Grid-mask. (d) The image generated from two samples are mixed according to a certain proportion. }\label{fig:auto}
\end{figure}

\section{Methods}
\subsection{Data Augmentation}
There are a total of 10789 images in the training and validation subset of the COCO2017 dataset, but the 80 categories with the background removed to show a long-tailed distribution. For example, there are more than 13000 person in the training set.
To effectively solve the imbalance of training data, we adopt some data augmentation strategies, such as bbox-jitter, grid-mask \cite{gridmask}, and mix-up \cite{mixup}, as shown in Fig. \ref{fig:auto}.

Bbox-jitter: we find that the number of boxes in the training sample is very pool.
To increase the fitting ability of the model, we increase the difficulty of the training samples.
It randomly shakes the boxes with an absolute amplitude to make the model more sensitive to the ROI area.
The bbox-jitter picture is present in Fig. \ref{fig:auto}(b).

Grid-mask \cite{gridmask}: It is an information dropping method based on the deletion of regions of the input image.
Given an input image, the grid-mask algorithm  randomly  removes a region with a disconnected pixel set, as shown in Fig. \ref{fig:auto}(c).
It can be found that the deletion region is neither a continuous region nor random pixels in dropout.
By deleting uniformly distributed areas, our model can achieve significant improvements in object detection.

Mix-up \cite{mixup}: The general data augmentation method is to transform the same category. A mix-up uses modeling between different categories to achieve data augmentation.
Two samples are randomly selected to improve the diversity of training set from the training samples for simple random weighted summation.
At the same time, the labels of the samples also correspond to the weighted summation, and then the prediction results and the weighted summation of the label after the loss are calculated, and the parameters are updated in the back propagation.
The resulting image for the mix-up is shown in Fig. \ref{fig:auto}(d).

\begin{figure}[tp]
	\centering
	\includegraphics[width=.9\linewidth]{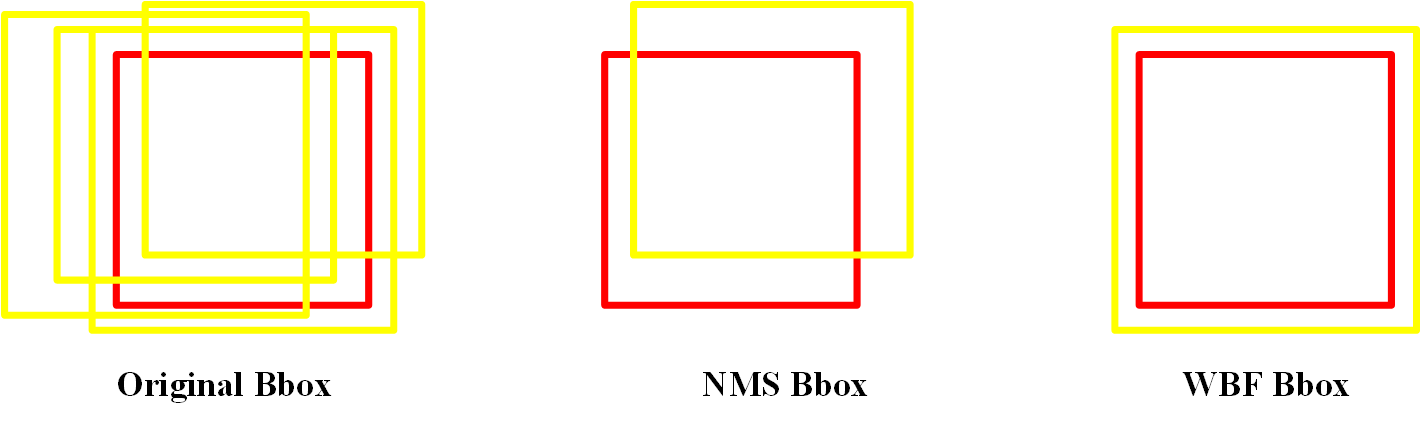}
    \vspace{-0.3cm}
	\caption{A simple comparison of  non-maximum suppression (NMS) and weighted boxes fusion (WBF).}\label{fig:nms}

\end{figure}

Besides the above three specific data augmentation methods, some regular augment methods are applied, such as random brightness contrast, random scale, pixel padding, random flipping, etc.
%In addition, we introduced two improvement techniques to obtain better performance (i.e., global context \cite{resnet} and weighted boxes fusion \cite{wbf}).

\subsection{Capturing Global Context Feature}
Comparing the performance of multiple detectors, we use a two-stage detector Cascade-RCNN and selected multiple different backbone networks, such as Res2Net-152 \cite{res2net}, SeNet-154 \cite{senet}, and ResNeSt-152 \cite{resnest}.
For the Cascade-RCNN detector,  the scenes' multiple salient targets, foreground interference, and chaotic backgrounds are challenging because the salient parts of different objects are individuated.  There is a lack of a global semantic relationship between multiple salient objects. In the usual feature pyramid networks (FPN) stage, the RoI features are usually extracted from a certain pyramid according to the RoI scale. This feature aggregation method of pyramids will be gradually diluted in high-level features passing to the bottom layer.

To address this problem, we recommend capturing the global context \cite{resnet} feature embedded into each ROI feature.
Specifically, we use the feature map of the entire image as a bounding box to achieve ROI convergence to obtain global context features.
Then, this global feature is tightly connected with the original feature of each region of interest in each stage.
Finally, the merged features are passed to the classification layer and box regression layer.

\subsection{Weighted Boxes Fusion}
Many model fusion methods can significantly improve performance, such as voting/Averaging, boosting, bagging, and stacking.
In this report, we adopt a weighted boxes fusion (WBF) \cite{wbf} integration algorithm that improves detection performance by integrating predictions of different object detection models.
For specific algorithm details, please refer to \cite{wbf}.
Here, we briefly introduce the difference between the weighted boxes fusion (WBF) and non-maximum suppression (NMS) algorithms.
The purpose of the NMS method is to exclude some frames, but the goal of WBF is to fuse the information of all prediction frames.
It can correct a situation in which all models predict the frame to be inaccurate, as shown in Fig. \ref{fig:nms}.
However, in this case, NMS wisdom leaves an inaccurate box. {\color{black}In contrast}, WBF uses information from three boxes to repair it.
We finally select the Cascade-RCNN networks of the three different backbone (i.e., Res2Net-152 \cite{res2net}, SeNet-154 \cite{senet}, and ResNeSt-152 \cite{resnest}) for an integrated prediction test dataset.

Besides, the above two specific improvement methods, we also do some special operations for Cascade-RCNN of ResNeSt-152.
We replace all batch normalization layers of the detection network with group normalization to make the statistical information of the normalization parameters more accurate.
Inspired by DetectoRS \cite{detectors}, we added switchable atrous convolution (SAC) to the backbone network.
Moreover, for Cascade-RCNN of three different backbone networks, we all use bounding box augmentation \footnote{https://github.com/mukopikmin/bounding-box-augmentation} to randomly crop and flip 6 times for categories with few samples.

\begin{table}[tp]
	\centering
	\caption{
		 Comparison results with Cascade-RCNN on validation set of COCO2017 subset.
	}\label{tab:val}
	\setlength{\tabcolsep}{1.5pt}
	\begin{tabular}{ccccc}
        \hline\noalign{\smallskip}
		\multicolumn{2}{c} {Methods}                           &AP@0.50:0.95 &AP@0.50 &AP@0.75  \\
        \noalign{\smallskip}
        \hline
        \noalign{\smallskip}
        &Res2Net-152					&32.5 		&52.7 &36.8\\
		&+Bbox-jitter					&33.3 		&53.1 &37.2\\
		&+Grid-mask  					&34.6 		&54.4 &38.4 \\
		&+Mix-up        				    &34.9   	&54.8 &38.7\\
		&+Global Context         		&36.3       &56.1 &40.2 \\	
		&SeNet-154 (Ensemble above strategy)        		&36.5       &56.6 &40.4 \\
		&ResNeSt-152  (Ensemble above strategy)      		&36.9       &56.4 &40.7 \\
\hline
	\end{tabular}
\end{table}

\begin{table}[tp]
	\centering
	\caption{
		 Comparison results with weighted boxes fusion (WBF) on test set of COCO2017 subset.
	}\label{tab:test}
	\setlength{\tabcolsep}{1.5pt}
	\begin{tabular*}{\hsize}{@{}@{\extracolsep{\fill}}ccccc@{}}
        \hline\noalign{\smallskip}
		&Methods                           &AP@0.50:0.95 &AP@0.50 &AP@0.75  \\
        \noalign{\smallskip}
        \hline
        \noalign{\smallskip}
        &Zuminternet        		&21.7       &34.7 &23.0 \\
		&AIGloud         		&24.0       &37.2 &25.4 \\
		&Rill        				    &33.0   	&49.2 &35.5\\
		&DeepBlueAI  					&35.1 		&53.0 &39.2 \\
		&JL					      &36.6 		&52.9 &40.0\\
        &Ours				&39.4 		&56.3 &42.7\\	
        \hline	
	\end{tabular*}
\end{table}

\section{Experiments}
In this section, we conduct several experiments on a subset of the COCO2017 detection dataset.
And the comparisons of detection performance are presented to verify the effectiveness of the proposed method and noted that we replace the original activation functions of RPN head and ROI head from the rectified linear unit (ReLU) \cite{relu} to mish \cite{mish}.

\subsection{Implement Detail}
All experiments are conducted using the mmdetection \cite{mmdetection} toolbox developed by PyTorch \cite{pytorch}.
And we run experiments on a NVIDIA V100 GPU.
Training configurations are summarized as follows.
(1)The input images are randomly sized to $4096 \times 1200$ and $4096 \times 1600$.
(2)For both mix-up and random flip operations, the implementation probability is set to 0.5.
(3)The amplitude of the bbox-jitter setting is between 0.95-1.05 times.
(4)The mini-batch stochastic gradient descent (SGD) method \cite{alexnet} is applied to optimize parameters. The weight decays are set to 5$\times$10$^{-4}$, and the momentums are set to 0.9.
(5) Because of the train from scratch and grid-mask operation, the maximum epoch for training is set to 144 (12 times \cite{rethinking}), and the batch size is 2.
(6) The initial learning is fixed to 0.02. Then, it decays to 0.002 at epoch 138 and 0.0002 at 142 epoch.
(7) On the original training and validation set, we first train on the category with fewer samples, and then train on the category with more samples. Finally, we adopt fine-tune on the expanded data set after bounding box augmentation strategy.
\subsection{Comparison of Different Strategies}
In the test phase, we use multi-scale ($4096 \times 1200$, $4096 \times 1400$ and $4096 \times 1600$) and randomly flipped fusion as the output result of a single model.
We evaluate different strategies on the validation set in Table \ref{tab:val}.
The baseline Cascade-RCNN with Res2Net-152 backbone network can achieves 32.5\% AP@0.50:0.95, 52.7\% AP@0.50 and 36.8\% AP@0.75 on validation set, respectively.
Besides, it can be found that all four strategies (i.e., bbox-jitter, grid-mask, mix-up and global context) can significantly improve the performance of the validation set. For example, with the bbox-jitter method, the AP@0.50:0.95 is 33.3\% higher than that of baseline.
With an ensemble of all the methods, the AP@0.50:0.95 is further improved by 3.8\% on the validation set.
Moreover, with the weighted boxes fusion (WBF) to refinement the prediction box,  our method achieves 39.4\% in the AP@0.50:0.95 score from Table \ref{tab:test}

\section{Conclusions}
This report details the key technologies used in the VIPriors object detection challenge. Our primary concern is the data augmentation to extract more compelling features. The introduction of bbox-jitter, grid-mask, and mix-up techniques to expand the training set make the model more robust. Besides, global context features are embedded in each region of interest (ROI) to help the model extract more significant object features. Finally, we adopt to a weighted boxes fusion (WBF) method of the multiple-model ensemble on the test dataset to modify the prediction box. Extensive experiments on a subset of the COCO2017 dataset demonstrate that the proposed method can obtain a competitive performance.

\ifCLASSOPTIONcaptionsoff
  \newpage
\fi

\bibliographystyle{IEEEtran}
\small{
\bibliography{reffullv2}
}

\end{document}